# Railway Network Delay Evolution: A Heterogeneous Graph Neural Network Approach

Zhongcan Li [a,b], Ping Huang [a,c], Chao Wen [a,d], Filipe Rodrigues [b]

[a] National Engineering Laboratory of Integrated Transportation Big Data Application Technology, Southwest Jiaotong University, Chengdu, 610031, China

[b] Technical University of Denmark (DTU), Kgs. Lyngby 2800, Denmark

[c] Institute for Transport Planning and Systems, ETH Zurich, Zurich 8093, Switzerland

[d] High-speed Railway Research Centre, University of Waterloo, Waterloo, N2L3G1, Canada

## Abstract

Railway operations involve different types of entities (stations, trains, etc.), making the existing graph/network models with homogenous nodes (i.e., the same kind of nodes) incapable of capturing the interactions between the entities. This paper aims to develop a heterogeneous graph neural network (HetGNN) model, which can address different types of nodes (i.e., heterogeneous nodes), to investigate the train delay evolution on railway networks. To this end, a graph architecture combining the HetGNN model and the GraphSAGE homogeneous GNN (HomoGNN), called SAGE-Het, is proposed. The aim is to capture the interactions between trains, trains and stations, and stations and other stations on delay evolution based on different edges. In contrast to the traditional methods that require the inputs to have constant dimensions (e.g., in rectangular or grid-like arrays) or only allow homogeneous nodes in the graph, SAGE-Het allows for flexible inputs and heterogeneous nodes. The data from two sub-networks of the China railway network, namely the Guangzhou South network (GZS-Net) and the Changsha South network (CSS-Net), are applied to test the performance and robustness of the proposed SAGE-Het model. The experimental results show that SAGE-Het exhibits better performance than the existing delay prediction methods and some advanced HetGNNs used for other prediction tasks; the predictive performances of SAGE-Het under different prediction time horizons (10/20/30 min ahead) all outperform other baseline methods; the accuracies are over 90% under the permissible 3-minute errors for the three prediction time horizons. Specifically, the influences of train interactions on delay propagation are investigated based on the proposed model. The results show that train interactions become subtle when the train headways increase (e.g., when the train headways are over 20 min, canceling the edges does not decrease the prediction performance). This finding directly contributes to decision-making in the situation where conflict-resolution or train-canceling actions are needed.

**Keywords:** Heterogeneous graph neural network, GraphSAGE, Railway network, Delay evolution; Train interactions

# 1. Introduction

Railway networks are complex systems consisting of multitudinous fixed facilities, including stations and tracks, and moving objects, most notably trains. The states of running trains are a result of interactions between the facilities and objects in the systems, as well as the effects of the external environment, such as bad weather. (Huang et al., 2020b; Wang and Zhang, 2019). As a measurement of the deviation between the actual and scheduled operation plans, delays are generally used to describe the state of a railway network, where trains are the most common moving objects in the system. As such, the delays of running trains are great indicators to evaluate the state of a railway network. Network-oriented train delay evolution is thus critical for railway operators and controllers as a comprehensive understanding of train delays in the network improves the quality of traffic control actions and rescheduling strategies (Sciences, 2018; Wen et al., 2019).

Due to the availability of data, numerous state-of-the-art machine-learning methods have been used in railway systems (Oneto et al., 2017; Ye et al., 2021). Among them, delay prediction or propagation is a popular field. However, the majority of previous delay prediction and propagation research is train-oriented (Huang et al., 2020b; Marković et al., 2015); in other words, they concentrated mainly on trains and forecasted the delays of each train in the downstream stations. However, in practice, dispatchers need to pay more attention to the network states from a systematic perspective, i.e., the delays of the railway network. In this study, we model train delays from a network perspective, considering the different kinds of trains and stations in the systems. The proposed network-oriented approach considers all trains at each moment, and explore the evolution of railway network delay, by predicting the delays of the running trains in the network after a given time interval. The network-oriented approach is expected to support dispatchers with a more comprehensive understanding of the state of the entire railway network, thus enabling them to make a global, rather than partial, adjustment plan.

Although all elements, such as trains, stations, disturbance events, etc., can be viewed as entities that interact to cause delays in the railway network, not all trains are simultaneously impacted by the same entities; for instance, at a given timestamp, some trains in the railway network are affected by a facility failure, but others are not. Thus, trains will have diverse input features in each prediction task. Previous studies were mainly based on traditional machine learning algorithms (e.g., the random forest (RF) (Nair et al., 2019), support vector regression (SVR) (Marković et al., 2015), etc.), graph-based approaches (e.g., graph neural networks (GNNs) (Heglund et al., 2020; Zhang et al., 2021), and Bayesian networks (Corman and Kecman, 2018; Lessan et al., 2019)). Machine learning models typically take rectangular or grid-like arrays as inputs (Sanchez-Lengeling et al., 2021); in other words, they require all samples to have the same input dimensions (i.e., the width and length of the array). To this end, some samples with missing features or values may be intentionally filled by human-specified values (e.g., zero or one). This artificial supplementation inevitably requires prior and domain knowledge, potentially increasing the model complexity and lowering the accuracy of results. The existing graph-based approaches for delay prediction only allow for homogeneous nodes (i.e., the same type of node). Heterogeneous GNNs (HetGNNs) update node information based on the adjacent edges and neighboring nodes, and allow nodes to have heterogeneous neighboring nodes (different kinds of neighboring nodes). Therefore, HetGNN is a great fit to address the complex systems with multiple types of entities to be considered.

The railway network in each timestamp can be viewed as a separate graph, with various entities (i.e., trains and stations) acting as nodes. The HetGNN can address heterogeneous neighboring nodes, and enable each node to connect with various numbers of nodes by edges (Hu et al., 2020; Wang et al., 2019). This means that HetGNN can eliminate the inconsistency of the input dimensions between train and station features. Thus, trains and stations can be viewed as different types of nodes. Then, connecting the trains to the interacting stations (or between trains), the HetGNN can be applied to explore the delay evolution of the railway network by predicting the delays of running trains on the graph (i.e., the railway network). We develop the delay evolution model (SAGE-Het) by using GraphSAGE (Hamilton et al., 2017) as the basic HomoGNNs in the proposed HetGNN model. Leveraging the latter, the strengths of interactions between nodes (i.e., trains and stations) are clarified by experimenting under different train headways. Thus, the results directly contribute to decision-making in the situation where traffic control/rescheduling action is needed. Therefore, the main contributions of this research are four-fold.

(1) A novel network-oriented HetGNN approach is proposed to predict the future delays of running trains on the whole network, which contrasts with existing models that only predict the delay of one train at downstream stations.

(2) A hybrid framework that can address heterogeneous nodes is developed to consider the interaction between different types of entities (trains and stations) in railway systems.

(3) A graph model called SAGE-Het, whose nodes allow for different numbers of heterogeneous neighboring nodes and flexible input, is put forward. This provides a better fit for complex systems with diverse numbers of influencing factors at different moments, such as railways.

(4) The SAGE-Het is leveraged to clarify the strength of train interactions by experimenting (canceling edges between train-train nodes) under different train headways (e.g., 3, 5, 10, and 20 minutes).

The remainder of this paper is organized as follows. In Section 2, related studies that focused on delay prediction and some applications of GNNs are reviewed. Section 3 describes the problem. Next, the method of this study, namely SAGE-Het, which combines the HetGNN model and the GraphSAGE model, is designed in Section 4. The data was introduced, and SAGE-Het model is implemented in Section 5. In Section 6, the performances of different baselines and SAGE-Het are compared and analyzed, and the train-train interaction on delay evolution is investigated. Finally, the conclusions and discussion of this work are presented in Section 7.

## 2. Literature Review

Delay prediction is a typical supervised-learning-based task, which usually aims at estimating the delay duration (Marković et al., 2015; Nair et al., 2019), delay influences (Huang et al., 2020a; Kecman and Goverde, 2015), and delay patterns (Huang et al., 2022). With the development of data collection and restoration techniques and computing ability, data-driven methods are widely employed in delay prediction based on historical data. According to the methods employed, related studies can be classified into those using traditional machine learning methods and GNNs.

The train operation process is a chronological process in which the train arrivals and departures can be abstracted as nodes, while the section running and dwelling times are treated as edges to link these nodes. Relying on this abstraction perspective, some Markov property-based methods are executed to predict arrival or departure delays. The Markov chain (MC) is an essential Markov property-based method that considers the last states to decide the current states. For delay prediction, MC-based studies usually consider that the arrival delays depend on the departure delay at the last station, while the departure delays are influenced by the arrival delays at the same stations (Barta et al., 2012; Gaurav and Srivastava, 2018; Kecman et al., 2015; Şahin, 2017). Another Markov property-based method is the Bayesian network (BN) . Compared with the MC, the BN can usually link several previous states in the network format (e.g., the current arrival can be linked with the arrival and departure states at the previous station(s)). In addition, the linking methods of the BN have been studied by relying on some optimization algorithms (Huang et al., 2020a; Lessan et al., 2019), resulting in hybrid BNs with better performance. Existing BN-based delay prediction studies have been applied to tackle several different problems, such as the delay lengths of trains (Corman and Kecman, 2018; Lessan et al., 2019; Li et al., 2021a), the delay effects (i.e., primary delays, the number of delayed trains, and the total delay times) (Huang et al., 2020a), the impacts of delays at stops on the network (Ulak et al., 2020), and the disruption duration (Zilko et al., 2016). These Markov property-based methods are highly interpretable to train operations, so they have been widely used, but these methods only considered train delays as the node attributes. However, delays are influenced by various internal and external factors of the railway system, meaning that the performance of these methods is hindered by this limitation.

Traditional machine learning methods used in delay prediction mainly include some machine learning methods, such as tree-based algorithms (e.g., the RF), SVR, artificial neural networks (ANNs), and hybrid neural networks. The decision tree is a classical machine learning method with a simple structure but high prediction accuracy [25]. Therefore, its variants have been widely employed. For example, the RF (Gaurav and Srivastava, 2018; Jiang et al., 2019; Klumpenhouwer and Shalaby, 2022; Nair et al., 2019), an ensemble algorithm based on decision trees, and some boosting algorithms (e.g., XGBoost (Li et al., 2020a; Shi et al., 2021), the gradient boosted decision tree (GBDT) (Wang and Zhang, 2019) have been used for delay prediction. The SVR is also another commonly used algorithm to model the delay propagation, such as delay time prediction (Barbour et al., 2018; Marković et al., 2015), delay recovery prediction (Wang et al., 2021), and running time prediction (Huang et al., 2020c). In addition, the $k$-nearest neighbors (KNN) algorithm (Li et al., 2016; Pongnumkul et al., 2014) and deep extreme learning machines (DELMs) (Oneto et al., 2017, 2018) are also applied for delay prediction. Simple ANNs have been applied to delay prediction in several railway networks (Kecman and Goverde, 2015; Peters et al., 2005; Yaghini et al., 2013), but with the development of neural network techniques, more advanced neural networks have been proposed to capture specific dependencies or address specific data attributes in railway systems. For instance, the recurrent neural network (RNN) and its variants (e.g., long short-term memory (LSTM) and gated recurrent units (GRUs)) are capable of handling time-series data. Train operation along stations and the chronological train arrival process at the station can both be treated as time-series processes; thus, LSTM has been used to extract the hidden information in the delay prediction process (Huang et al., 2020b; Li et al., 2022; Wen et al., 2020). However, the train operation process is influenced by different kinds of entities, which makes it difficult to make an accurate prediction via the use of only a single neural network. Therefore, hybrid neural networks that can use diverse neural network blocks to handle different kinds of data, such as the combinations of LSTM and Fully-connected neural network (FCNN) (Huang et al., 2020b), a convolutional neural network (CNN) and FCNN (Huang et al., 2021; Huang et al., 2020d), and a CNN, LSTM, and FCNN (Li et al., 2022), have been proposed to predict delays. These traditional machine learning methods are typical supervised learning methods by which the relationship between the input and output features are learned, based on which the future is then predicted. In the modeling process of these traditional machine learning methods, the influencing factors (i.e., input features) for each sample must be the same, which requires samples with some missing input features to be filled artificially. This artificial supplementation inevitably requires prior and domain knowledge, potentially increasing the model's complexity and lowering the accuracy of results.

GNN-related studies have achieved numerous successes, thus contributing to their wide applications in many transportation fields, such as traffic state prediction (Cui et al., 2019), traffic flow prediction (Guo et al., 2019; Wang et al., 2020), and traffic demand prediction (Zou et al., 2021). In terms of railway delay prediction, some studies (Heglund et al., 2020; Li et al., 2021b) have applied graph convolutional networks (GCNs) with a spatiotemporal attention mechanism to predict railway delays. Similarly, based on the spatiotemporal GCN (STGCN), a previous study (Zhang et al., 2021) predicted the number of delayed trains at each station in the railway network by taking stations as nodes. In addition, a multi-layer time-series graph neural network (MTGNN) model has been established to predict the arrival delay under diverse delay causes (Ding et al., 2021). Alternatively, by considering each train at each previous station (Ding et al., 2021; Li et al., 2021b) or section (Heglund et al., 2020) as a node, the delay(s) of a given train at the downstream station(s) were estimated according to the delay(s) at the previous station(s). As with traditional machine learning studies, these HomoGNN-based delay prediction studies were single-train-oriented (Ding et al., 2021; Heglund et al., 2020; Li et al., 2021b). These railway delay prediction studies all used HomoGNNs, which can only address homogenous nodes. Suppose some samples have specific influencing factors while others do not (e.g., some delayed trains are influenced by a facility failure, while others do not). In this case, these samples must be filled artificially or deleted. HetGNNs can address the shortcomings of the previous HomoGNN-based delay prediction studies, because they can have heterogeneous nodes. HetGNNs have been used in text classification (Linmei et al., 2019), system recommendation (Fan et al., 2019), and spam review detection (Li et al., 2019). In the traffic domain, an approach based on the heterogeneous graph attention network for the prediction of traffic speed has been proposed (Jin et al., 2021). The trajectories of different traffic participants (e.g., vehicles, pedestrians, and cyclists) have been predicted using a hierarchical HetGNN model (Li et al., 2021c). In terms of railway delay prediction, there is no HetGNN-based study.

The preceding review of the existing delay prediction literature demonstrates that previous studies mainly relied on various traditional machine learning or HomoGNN methods. These methods require rectangular or grid-like arrays as input features. If the samples have different numbers of features, the values must be filled artificially. In other words, previous studies could only address consistent input features (with the same input dimensions). HetGNN updates the information of nodes by relying on their neighboring nodes (edges), and each node can have different numbers of heterogeneous neighboring nodes (edges). Thus, to address the gap, this paper proposes a HetGNN model in which trains and stations are treated as nodes and linked with the corresponding affected train nodes to explore railway delay evolution. This allows the model to take the interactions between different types of trains and stations in railway systems into account, thus potentially boosting the accuracy of delay evolution estimation.

## 3. Problem Statement

Railway systems are composed of various entities, including stations, trains, tracks, extraneous environments, etc. Train delays, which are the consequences of interactions among diverse entities in the railway network, are mainly used to measure the state of the railway network. We propose a network-oriented approach to explore the evolution of railway network delay by estimating the future delays of running trains, considering the influences of diverse entities. Let us use **Fig. 1** to clarify the delay evolution problem. In the figure, TS 1 represents a terminal station, which has multiple arrival-departure yards to serve train operations. PS 1 to PS 11 represent passing (intermediate) stations, which are connected only by one railway line. **Fig. 1(A)** presents the current railway network state (at timestamp $T$), while the to-be-predicted railway network state (at timestamp $T + \Delta T$) is exhibited in **Fig. 1(B)**. We assume that Trains 0-7 are running from passing stations toward the terminal station TS 1 at timestamp $T$; therefore, they are running trains; Train 8 has already arrived at TS 1; it is, therefore, called the terminated train at timestamp $T$. TS 1 has three arrival-departure yards (TS 1-1, TS 1-2, TS 1-3 in **Fig. 1**) to serve trains on four railway lines. These yards are linked with each other. The aim of this work is to estimate the delays of all running trains at $T+\Delta T$ on a network, based on the current railway network state (i.e., at timestamp $T$). We model the effects and interactions between trains and stations using a heterogeneous graph $G$, where $G = (V, E)$ ($V$ and $E$ denote the node and edge sets, respectively). We construct the graph $G$ trying to model the following influences between (of) entities in the railway system.

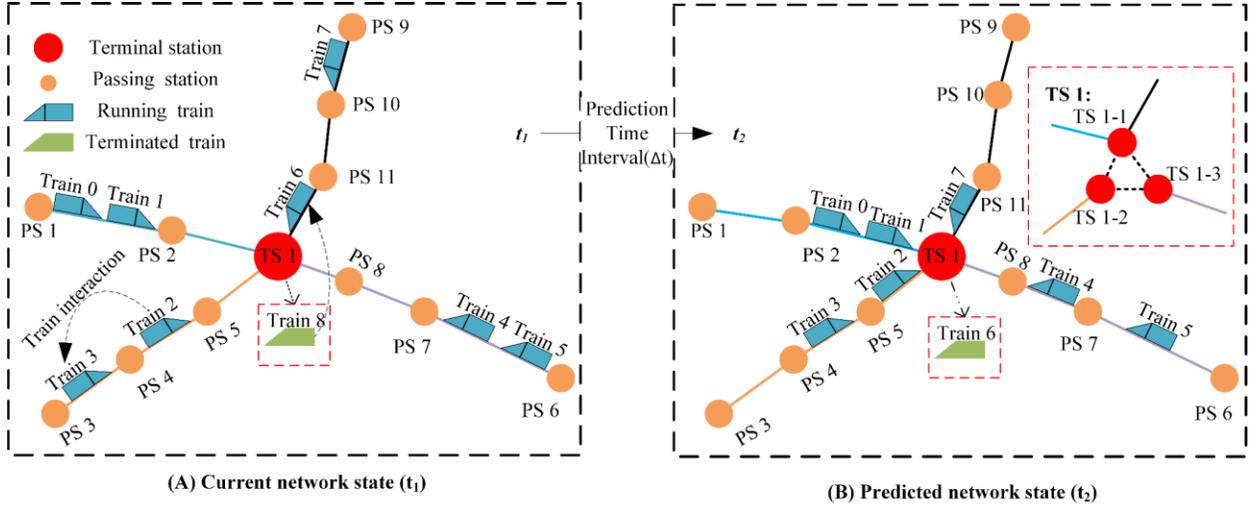

**Fig. 1.** The railway network delay evolution process.

First, train-train interactions are crucial for delay evolution; for example, in **Fig. 1,** if Train 1 is delayed and the headway between it and Train 0 is short, Train 0 may be delayed due to the minimum headway requirement. We use the change of running trains' delays to explore the railway delay evolution on the network. Given a particular timestamp, running trains differ from terminated trains because terminated trains have already arrived at terminal stations, meaning that terminated trains do not need to be predicted in the next timestamp. However, terminated trains will influence other running trains. This means that all trains should be addressed as nodes in the graph to capture the train interactions. Therefore, running trains and terminated trains are defined as RT and TT nodes in the heterogeneous graph (i.e., the railway network). Edges (also known as metapaths) reflect the interactions between nodes. We define two kinds of edges to reflect the train-train interaction, including the interactions between RTs and between RTs and TTs. Here, the edges between RTs are designated as *RT-rr-RT*, and the edges between RTs and TTs are defined as *RT-rt-TT*. The relationship between nodes is represented by the intermediate abbreviations (e.g., *rr* and *rt*) in the edges; for instance, *write* is the relationship of node Author and Paper in the edge *Author-write-Paper*.

Second, trains update their delay states when passing the signal record points, which are the stations in this study (i.e., trains update the delay states at stations). Thus, stations also influence the delay evolution of the network. A typical example is that trains have to wait for the opening of the train routes (releasing of the block) before entering the station due to potential route conflicts. From the perspective of railway operations, there are two types of stations: passing (intermediate) and terminal stations. Trains only pass through passing stations while they terminate at terminal stations. In addition, Passing stations are usually traversed by only one railway line (as shown in **Fig. 1**, where there is only one railway line from PS 1 to PS 2). Therefore, only trains running on one line will arrive at or depart for passing stations. However, the terminal station may have several arrival-departure yards linking different railway lines (e.g., TS 1 has TS 1-1, TS 1-2, and TS 1-3 in **Fig. 1**). Trains running on multiple lines may arrive at or depart from TSs. For example, trains on any of the four railway lines can arrive at or depart from TS 1 in **Fig. 1**. In TSs, trains may run from one yard to another for transferring lines.

This means that interactions between trains from different lines at terminal stations are possible. Hence, passing stations and the arrival-departure yards of terminal stations are referred to as PS and TS nodes, respectively, in the proposed graph model, meaning that a terminal station may contain multiple nodes. TS and PS nodes are referred to as station nodes in this study. We name the edges between station nodes and RT nodes *Station-sr-RT* and the edges between stations nodes *Station-ss-Station*.

The proposed graph, thus, contains four types of nodes, namely RTs (e.g., Train 0 in **Fig. 1(A)**), TTs (e.g., Train 8 in **Fig. 1(A)**), PSs (e.g., PS 1 in **Fig. 1(A)**), and TSs (e.g., TS 1-1 in **Fig. 1(A)**). In addition, there are four types of edges to reflect the interaction between nodes, including *RT-rr-RT, RT-rt-TT, Station-sr-RT, and Station-ss-Station*.

Thus, the aforementioned delay evolution problem on a railway network can be described by:

$$F([V^T, E^T]; G) = V_1^{T+\Delta T} \quad , \tag{1}$$

where $F()$ is the HetGNN model, $\Delta T$ is the prediction time interval/horizon, $V^T$ and $E^T$ are the node and edge sets at timestamp $T$. Additionally, $V^T = \{V_1^T, V_2^T, V_3^T, V_4^T\}$, where $V_1^T, V_2^T, V_3^T, V_4^T$ denotes RT, TT, PS, and TS nodes at timestamp $T$, respectively. $E^T = \{E_1^T, E_2^T, E_3^T, E_4^T\}$ are the edge sets, where $E_1^T, E_2^T, E_3^T, E_4^T$ represent *RT-rr-RT, RT-rt-TT, Station-sr-RT,* and *Station-ss-Station* edges at timestamp $T$, respectively. It is noted that we only consider the trains running to terminal stations on the network; therefore, downstream stations will affect the upstream stations, but not the other way around. For instance, only PS 2 influence train operations at PS 1. In other words, graph $G$ is a directed graph.

## 4. Method

### 4.1 The procedure of the proposed method

To predict the delays of all running trains on a railway network, a HetGNN-based approach is proposed in this study. The overview of the HetGNN approach is described in **Fig. 2**. The proposed approach contains four steps, i.e., Step (a) to Step (d). In Step (a), the railway network at timestamp $T$ is abstracted as a heterogeneous graph $G = (V^T, E^T)$ , and is used as the input of our HetGNN model. In Step (b), the nodes are updated by the Homogeneous Model or Heterogeneous Model (shown below **Fig. 2**), depending on whether there are one kind or multiple kinds of edge(s) connecting them. If multiple types of edges connect a node, the updated node result will be the aggregation relying on the updated results from each edge type. For instance, RT nodes will be updated relying on edges *RT-rr-RT, TT-rt-RT, Station-sr-RT*. The output of Step (b) is then aggregated in Step (c). Step (b) and Step (c) can be repeated several times, i.e., nodes are updated by several convolutional layers in HetGNN. Finally, the delays of RT nodes at timestamp $T+\Delta T$ (i.e., $V_1^{T+\Delta T}$ ) are obtained in Step (d), by feeding the aggregated RT nodes into a linear layer.

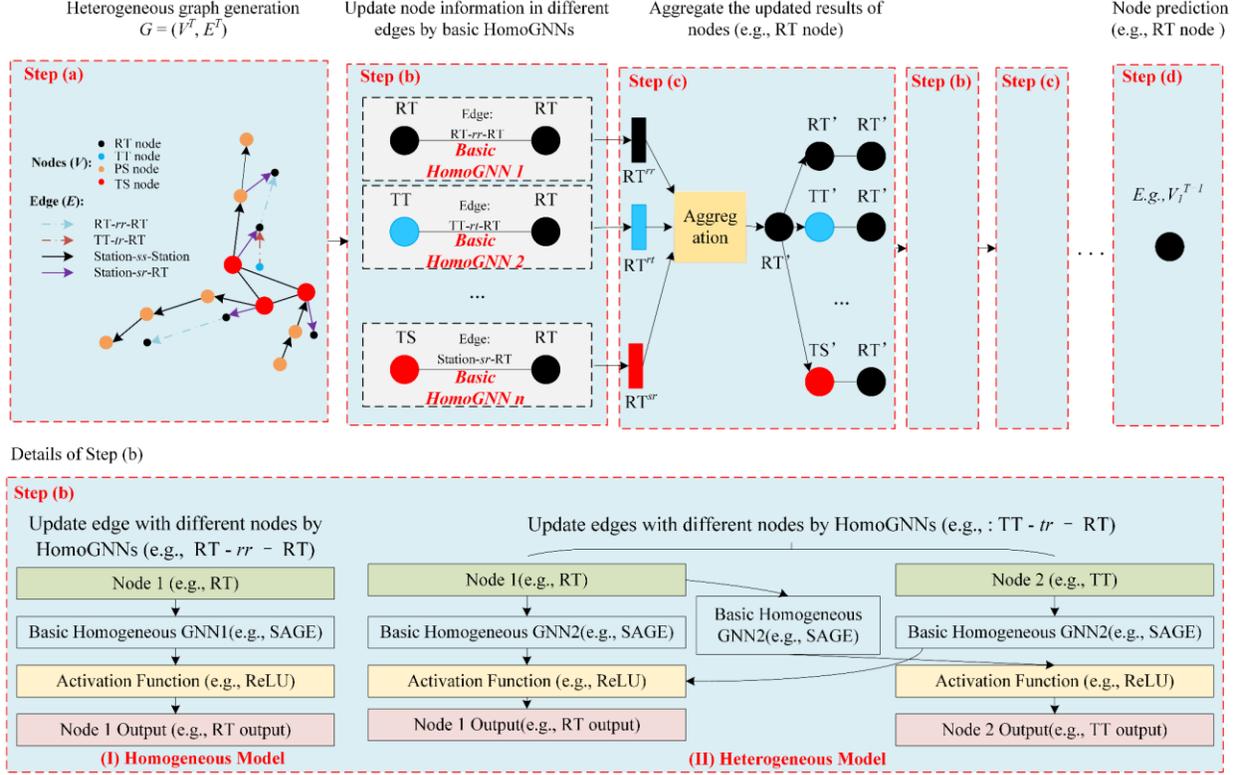

**Fig. 2.** Overview of the proposed HetGNN-based approach in this study.

### 4.2 Heterogeneous Graph Neural Network·(HetGNN)

The HetGNN·takes a heterogeneous graph as input. In this study, trains and stations are addressed as nodes to form heterogeneous graphs, and the HetGNN is used to process different types of entities in railway systems. Subsequently, the HetGNN updates node information based on the convolutional layers, resulting in an updated graph. Finally, the prediction task, such as node predictions or graph classification tasks, can be executed using the updated graph.

First, HetGNN·takes a heterogeneous graph as input. The heterogeneous graph contains different kinds of nodes, and the nodes connect each other based on diverse edges. For instance, the railway network heterogeneous graph ($G = (V, E)$) consists of nodes RT, TT, PS, and TS and edges *RT-rr-RT, TT-tr-RT, Station-sr-RT,* and *station-ss-station*.

Subsequently, nodes connecting with different edges will be updated based on the HetGNN convolutional layers. The Homogeneous Model and Heterogeneous Model (shown in **Fig.2 Step (b, I and II)**) are used to update node information in HetGNN convolutional layers. In heterogeneous graphs, nodes at the edges whose two nodes at both ends are homogeneous (e.g., the edge *RT-rr-RT*) can be updated by standard HomoGNNs in HetGNN convolutional layers. The node features updated by the Homogeneous Model for node $i$ in layer $k$, namely $x_i^{(k)}$, are updated by

$$x_i^{(k)} = r^{(k)} \left( x_i^{(k-1)}, \square_{j \in N(i)} \phi^{(k)} \left( x_i^{(k-1)}, x_j^{(k-1)}, e_{ji} \right) \right), \qquad (2)$$

where $e_{ji}$ denotes edge features from node $j$ to node $i$, $N(i)$ represents the neighboring node set of node $i$, and function $\square$ is a differentiable, permutation-invariant function (e.g., the sum, mean, or max). Moreover, $\phi$ and $r$ denote differentiable functions, which can include many existing excellent models, such as graph attention networks (GATs) (Veličković et al., 2017), GCNs (Kipf and Welling, 2016), and GraphSAGE (Hamilton et al., 2017).

For the edges connecting heterogeneous nodes (e.g., *TT-tr-RT* in our case), the nodes cannot be updated by the standard HomoGNNs. Therefore, the Heterogeneous Model is needed to update the node information. Since the feature dimensionalities of the nodes in the edge are different, this Heterogeneous Model first applies the basic HomoGNN (shown as the basic HomoGNN 2 in **Fig.2 Step (b, II)**) to map the two distinctive nodes into the same dimensional space. Subsequently, the node information updated by the basic HomoGNN is aggregated by a summation function, and the aggregated result is then activated by a ReLU activation function, and the destination nodes (e.g., the RT nodes in the edges *TT-tr-RT* in our case) are finally updated.

Based on the HetGNN convolutional layers, all nodes in different kinds of edges can be updated, resulting in an updated heterogeneous graph. The mathematical expression of the HetGNN convolutional layers for all nodes can be represented by Eq (3), as follows,

$$v_i^{(k)} = \bigoplus_{e \in E} f_\theta^{(k,e)} \left( v_i^{(k-1)}, \{ v_w^{(k-1)} : w \in N^{(e)}(i) \} \right), \qquad (3)$$

where $v_i^{(k)}$ is the node features of node $i$ in the $k_{th}$ layer, and the nodes connecting various edges are updated based on the basic HomoGNN $f_\theta^k$ in the $k_{th}$ layer. Additionally, $N^{(e)}(i)$ denotes the set of corresponding beginning nodes which point to the destination node (e.g., the TT node in *TT-tr-RT* edges when the updated node is RT) in the edge $e$, and the $\oplus$ is aggregation function (e.g., the sum, max, or min) for aggregating node information generated by different edges.

Lastly, based on the new heterogeneous graph updated by the HetGNN convolutional layers, the prediction task can be executed to predict the nodes or graph classification tasks.

### 4.3 GraphSAGE-based HetGNN Architecture

We employ GraphSAGE (Hamilton et al., 2017) as the basic HomoGNNs in the proposed model as it has achieved satisfactory performance in numerous prediction tasks. GCNs are the most commonly-used HomoGNNs in the existing GNN-based delay prediction studies. Compared with the classical GCN, GraphSAGE updates node information through random sampling of the neighboring nodes during training, which can control the number of nodes involved in the calculation, and therefore dramatically reduces training resources. In this way, GraphSAGE updates node information based on its neighbors without the need to re-iterate the entire graph when a node is added, making it have a better generalization ability than GCN. In other words, for all basic HomoGNNs in Step (b), the functions $\phi$ and $r$ in Eq (2), and $f_\theta^k$ in Eq (3) are GraphSAGE models in the proposed model. Thus, we refer to the proposed model as SAGE-Het. The number of convolutional layers in the SAGE-Het (i.e., repeated Step (b) and (c)) are crucial hyperparameters. After hyperparameter-tuning, the number of convolutional layers is selected as four, each with 256 neurons and a ReLU activation function. In addition, all aggregation functions (i.e., $\square$ in Eq. (2) and $\oplus$ in Eq. (3)) are performed by the summation functions.

A brief summary of the hyperparameters in SAGE-Het is provided in **Table 1**.

**Table 1.** Summary of the SAGE-Het hyperparameters

| | |
|---|---|
| Model input and output | **Input:** The heterogeneous graph $G = (V^T, E^T)$ , i.e., the railway network state at timestamp $T$ <br> **Output:** $V_i^{T+\Delta T}$ , i.e., the delays of RT nodes at timestamp $T+\Delta T$ |
| Basic HomoGNNs | GraphSAGE |
| Number of convolutional layers | 4 |
| Neurons in each convolutional layer | 256 |
| Aggregation Function | Sum |

Let us consider **Fig. 2** as an example to introduce the prediction process of delay evolution. The railway network state at timestamp $T$ (i.e., $G = (V^T, E^T)$) is first taken as the input. Steps (b) and (c) are the schematic diagram of SAGE-Het convolutional layers; therefore, they are used to update the node information in different edges. For the RT nodes in the edges *RT-rr-RT*, as both ends in the edges are homogeneous, they are updated by using the Homogeneous Model (shown in **Fig. 2 step (b, I)**). In this case, $x_i^{(k)}$ is the feature of the RT node $i$ in layer $k$, $N(i)$ represents the set of the beginning RT nodes in the edges *RT-rr-RT* (the first RT node) for RT node $i$, and the function $\phi$ and $r$ are the GraphSAGEs. The RT nodes in the edge *TT-rr-RT* and *Station-sr-RT* are updated based on the Heterogeneous Model (shown in **Fig. 2 Step (b, II)**). Then, the nodes information updated from different edges (i.e., $RT^{rr}$, $RT^{rt}$, and $RT^{sr}$) can be aggregated by a summation function (the "aggregation" block in Step (c)), and the RT nodes after aggregation are then obtained (e.g., RT' in **Fig. 2**). The same process can be applied to other nodes (such as TS nodes), thus producing a new heterogeneous graph updated by the SAGE-Het convolutional layer. Steps (b) and (c) can be repeated to represent that all nodes are updated via several successive SAGE-Het convolutional layers. It should be noted that each SAGE-Het convolutional layer can vary by using different basic HomoGNNs. Finally, the RT nodes are predicted by feeding the RT nodes in the new heterogeneous graph into a linear layer.

The pseudo-code of the SAGE-Het is shown in Algorithm 1.

**Algorithm 1**. The pseudo-code of the GraphSAGE HetGNN model

**Input:**

The heterogeneous graph $G = (V^T, E^T)$ ,i.e., the railway network state at timestamp $T$

Node type: $V^T = \{V_1{}^T, V_2{}^T, V_3{}^T, V_4{}^T\}$

Edge type: $E^T = \{E_1{}^T, E_2{}^T, E_3{}^T, E_4{}^T\}$

Node feature: $x_i, \forall i \in V$

Number of convolutional layers (i.e., Step (b) and Step (c) in **Fig.2**): Num

**Output:** $V_1{}^{T+\Delta T}$ (i.e., the delays of RT nodes at timestamp $T+\Delta T$)

---

**For** layer $k$ in Num:

    **For** edge $e$ in $E^T$ do:

        **For** node $i \in V^T$ in $e$:

            **If** nodes in $e$ are the same:

                Update the node information based on the Homogeneous Model with GraphSAGE basic HomoGNN (shown in **Fig.2 Step(b, I)**)

            **If** nodes in $e$ are different:

                Update the node information based on the Heterogeneous model with GraphSAGE basic HomoGNN (shown in **Fig.2 Step(b, II)**)

        Sum the updated results of the nodes in the edge $e$

        **End**

    Sum the results of all nodes in all edges updated by the $k_{th}$ convolutional layer.

    Then, obtain new edges and nodes, e.g., edges RT'- RT', TS'-RT', and TT'-RT' in **Fig.2** Step (c).

    **End**

All nodes updated after $k$ convolutional layers are obtained

**End**

RT nodes updated after $k$ convolutional layers are fed into a linear layer to predict the delays of RT nodes timestamp $T+\Delta T$ (i.e., **in Fig.2 Step(d)**).

**Return** $V_1{}^{T+\Delta T}$

---

# 5. Model Implementation

## 5.1 Data Description

    The data used in this study were obtained from the Guangzhou Railway Bureau, China, and the period of the data is from 03/24/2015 to 11/10/2016. Two railway networks, namely the Guangzhou South network (GZS-Net) and the Changsha South network (CSS-Net), were taken as cases to validate the proposed SAGE-Het model. The schematic diagrams of GZS-Net and CSS-Net are shown in **Fig. 3**. Guangzhou South (GZS) is the terminal station of five HSRs, namely the Jing-Guang (JG) HSR, Nan-Guang (NG) HSR, Gui-Guang (GG) HSR, Guang-Zhu (GZ) HSR, and Guang-Shen (GS) HSR. Changsha South (CSS) is also a multi-line station, which can be deemed the connection station of four HSRs, namely the Jing-Chang HSR, Kun-Chang HSR, Chang-Guang (CG) HSR, and Hu-Chang (HC) HSR.

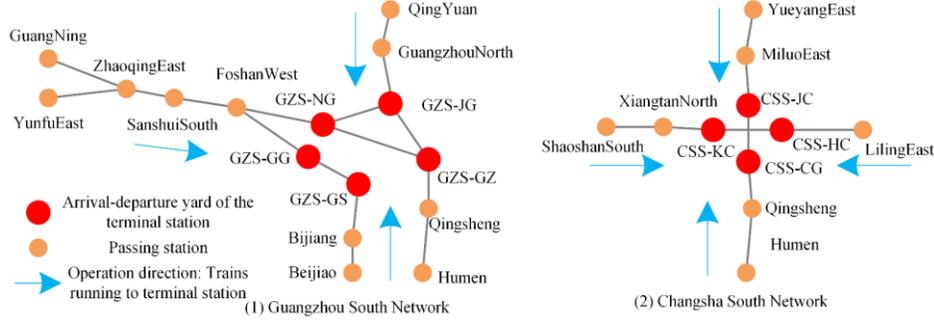

**Fig. 3.** The schematic diagram of the Guangzhou South network and the Changsha South network.

In this paper, the historical train operation data is used, and all data is recorded in minutes. Some samples of the raw operation data are shown in **Table 2**, including the Date, Train number, Station, Scheduled arrival, Scheduled departure, Actual arrival, Actual departure, and Occupied track. The attributes of train nodes (RT and TT nodes) are calculated and used to explore the delay evolution based on the raw data. In addition, the number of station tracks (the variable $N$) can be obtained from the corresponding station (yard) layouts. As the raw data only recorded the train operation states at stations, train delay states are updated when the trains arrive at and depart from the station.

**Table 2.** Some samples of the raw operation data

| Train Number | Station | Date | Actual arrival | Actual departure | Scheduled arrival | Scheduled departure | Occupied track |
|---|---|---|---|---|---|---|---|
| D903 | Guangzhou North | 2015/3/24 | 6:09 | 6:09 | 6:13 | 6:13 | I |
| G6023 | Hengshan West | 2016/1/5 | 14:13 | 14:15 | 14:09 | 14:11 | 3 |
| G6023 | Guangzhou South | 2016/1/5 | 16:30 | 16:33 | 13:30 | 16:35 | 11 |
| G6143 | Changsha South | 2015/4/4 | 12:03 | 12:19 | 11:38 | 11:52 | 3 |

Note: For the occupied track, the passage tracks at the station are labeled with Roman characters, while the dwelling tracks are labeled with numbers

As described in Section 3, this paper only considers the trains heading to the terminal station (e.g., towards TS 1 in **Fig.1**), while trains heading away from terminal stations are not considered. Railway networks at different timestamps can be abstracted to various heterogeneous graphs, as shown in **Fig. 4. Fig. 4(A)** is a heterogeneous graph sample, which is generated from the railway network at timestamp $T$ (i.e., it corresponds to **Fig.1 (A)**). Similarly, **Fig. 4(B)** is another sample that corresponds to **Fig.1 (B)**.

Taking **Fig. 4** as an example, the aim of this study is to predict the delays of the RT nodes of **Fig. 4(A)** at timestamp $T+\Delta T$, taking the heterogeneous graph sample in **Fig. 4(A)** (the heterogeneous graph at timestamp $T$) as the model input, with the prediction time interval $\Delta T$. It should be noted that the delays of terminated trains do not need to be predicted in the next timestamp, as they have already arrived at the terminal stations. Therefore, the graph structure at $T+\Delta T$ is different from that at $T$.

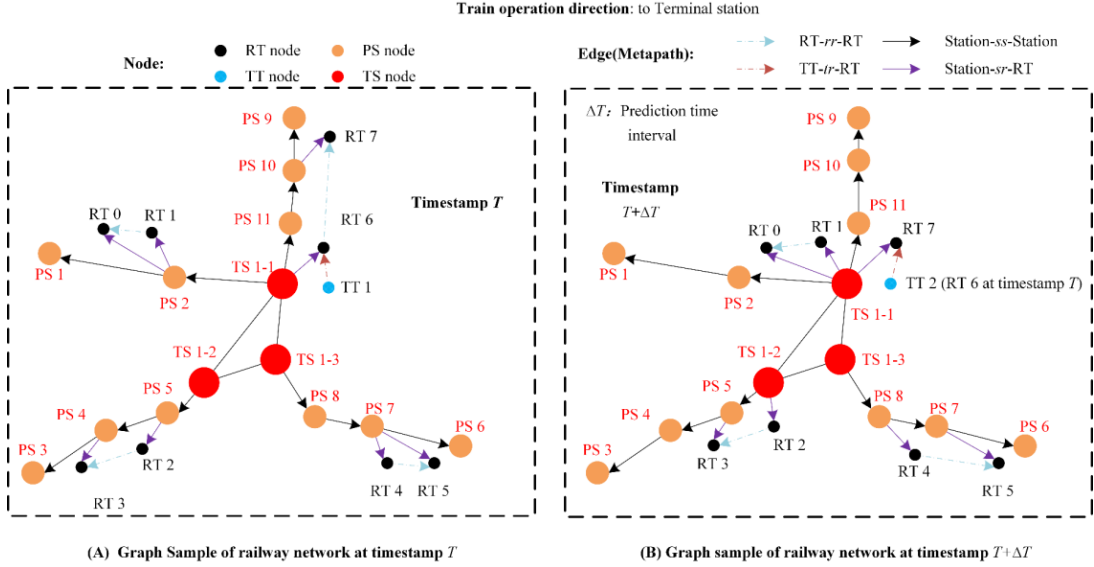

**Fig. 4.** The heterogeneous graphs abstracted from the railway network at different timestamps.

The graphs vary at different timestamps due to the differences in the number of running trains at various times. For example, **Fig. 4(A)** has 8 RT nodes, while **Fig. 4(B)** has 7 RT nodes. Railway networks at each timestamp can be abstracted to a heterogeneous graph sample, which is shown in the two graph samples in **Fig. 4(A)** and **(B)**. The prediction time interval ($\Delta T$) is an adjustable hyperparameter, which reflects how far ahead the railway network state is predicted. The prediction time interval of this paper was first set as 20 min, with reference to the *2018 RAS competition* (Sciences, 2018). Thus, the railway networks (graphs) were updated every 20 min, and the delays of running trains 20 min later were predicted based on the current graph structure. The prediction periods were only considered from 8:00 to 23:00 each day (i.e., timestamps should be 8:00, 8:20, …, 23:00), because too few (or no) trains run in the railway network during other periods. Graph samples with RT node delays more than 90 min were eliminated because the proportions of these delays were too small (less than 1% in GZS-Net and CSS-Net) to ensure that the model was correctly trained. This elimination guaranteed the statistical significance and sufficient training of the HetGNN model, and the considered horizon could be prolonged once there were enough of these delayed data (Corman and Kecman, 2018; Lessan et al., 2019).

The predicted delays of the RT nodes are unknown, so, if only the delayed trains are applied to train the model, in practical application, some early and on-time arrivals would also be predicted as delays (Corman and Kecman, 2018; Huang et al., 2020d; Lessan et al., 2019). Thus, each train could be assigned a value, corresponding to the actual arrival/departure minus scheduled arrival/departure, to describe the delay state. The value of an on-time arrival was 0, while early arrivals had negative values. The delays considered in this study include early arrivals, on-time arrivals, and delayed arrivals, which are imperative. Ultimately, the datasets of GZS-net and CSS-net had 23417 and 23879 graph samples respectively. These graph samples were shuffled, and 60% of the graph samples were used for model training, 20% were used as the validation dataset to tune the model hyperparameters, and the remaining 20% were employed for model testing.

*5.2 Node and Edge Feature Determination*

**(1) Features of RT nodes**

In this study, the network states (train delays) are predicted based on the graph information. The node and edge features first need to be determined to predict train delays. Because delays are usually (and are in this study) calculated on train arrival/departure events, we first define some train operation-related variables. The following variables are used to express the node features:

$A_{Train,Station}^{Sch}$ : the scheduled arrival time of the *Train* at the *Station*;

$D_{Train,Station}^{Sch}$ : the scheduled departure time of the *Train* at the *Station*;

$A_{Train,Station}^{Act}$ : the actual arrival time of the *Train* at the *Station*;

$D_{Train,Station}^{Act}$ : the actual departure time of the *Train* at the *Station*;

$AD_{Train,Station}$ : the arrival delay of the *Train* at the *Station*, $AD_{Train,Station} = A_{Train,Station}^{Act} - A_{Train,Station}^{Sch}$ ;

$DD_{Train,Station}$ : the departure delay of the *Train* at the *Station*, $DD_{Train,Station} = D_{Train,Station}^{Act} - D_{Train,Station}^{Sch}$ ;

The factors influencing delay prediction are determined by referencing previous delay prediction studies (Huang et al., 2020b; Li et al., 2020b; Nabian et al., 2019) and the variables defined above.

For the RT nodes, the current delays (*C*) are the basis of the subsequent delay state in delay evolution. Thus, current delays are considered as features of RT nodes. When a train (e.g., Train 1 in **Fig. 1**) is dwelling at a Passing station at timestamp *T*, *C* should be the arrival delay of the train (i.e., $AD_{Train1,Station1}$). If a train (Train 1 in **Fig. 1**) is running in section (section (*Station 1 Station 2*)) at current timestamp (timestamp *T*), then *C* should be the departure delay of the train at *Station 1* (which is $DD_{Train1,Station1}$). For instance, at timestamp *T*, the *C* of RT 0 is the departure delay of RT 0 at PS 1 in **Fig. 4(A)**.

The headway between the to-be-predicted trains and forward trains, labeled as *I*, reflect the train-train interaction. Due to the minimum headway requirement, if the headways are too small, the delayed forward train may influence the rear trains, resulting in knock-on delays. In contrast, when the headways are long enough, the rear trains will not be influenced by the delayed forward trains. Thus, *I* is another critical factor for delay evolution, and therefore is also considered as the attribute of RT nodes. For instance, at timestamp *T*, the *I* of RT 0 (i.e., Train 0 in **Fig. 1(A)**) is the headway between RT 1 (i.e., Train 1 in **Fig. 1(A)**) and RT 0 (equaling $D_{Train\,0,PS\,1}^{Act}$ - $D_{Train\,1,PS\,1}^{Act}$ ) in **Fig. 4(A)**.

The supplement time between the current timestamp (*T*) and the to-be-predicted timestamp (*T*+Δ*T*) is crucial for running trains as it determines the maximum recovery times. The supplement time is the difference between the scheduled running time and the minimum running time. We define the scheduled running time of RT nodes from timestamp *T* to timestamp *T*+Δ*T* as the variable *S*, while the minimum running time of RT nodes from timestamp $T$ to timestamp $T+\Delta T$ is the variable $M$. The variable $S$ is the scheduled running time in the section(s) where the trains pass through during $\Delta T$. For example, in **Fig. 1(A)**, Train 7 is running in the section (PS 9, PS 10) at timestamp $T$, and it will be running in the section (PS 11, TS 1) based on the scheduled timetable at timestamp $T+\Delta T$. Because trains update their delay states at stations, we will only know the scheduled delay state of Train 7 at PS 11. In this way, the variable $S$ is the scheduled running time of Train 7 from PS 9 to PS 11 during $\Delta T$, i.e., $S = D_{Train7,PS11}^{Sch} - D_{Train7,PS9}^{Sch}$. Similarly, the variable $M$ is the minimum running time in the section(s) where the trains pass through during $\Delta T$; for instance, the variable $M$ is the scheduled running time of Train 7 from PS 9 to PS 11 in the above case. Generally, $M$ is different from trains in the same section because of the diverse service types.

In addition, the scheduled remaining running times from now ($T$) to the the to-be-predicted timestamp ($T+\Delta T$) is also critical, and it is labeled as $R$. It should be noted that $R$ should be the scheduled running time from current timestamp ($T$) to the nearest delay updating timestamp after $\Delta T$, because trains update delay states at station in thi study. Taking **Fig. 1** as an example, Trains 4 and 5 run in the same section at timestamp $T$, while Train 4 runs in the downstream section and Train 5 still runs in the previous section (i.e., Train 4 updates its delay state but Train 5 does not) at timestamp $T+\Delta T$. For Train 4, the nearest delay updating timestamp after $\Delta T$ is the departure time at PS 7; therefore $R = D_{Train4,PS7}^{Sch} - T$. The variable $S$ for Train 4 can be represented as $S = D_{Train4,PS7}^{Sch} - D_{Train4,PS6}^{Act}$, so $R$ can also be $R = S - (T - D_{Train4,PS6}^{Act})$ for Train 4 in **Fig. 1**. Assuming that Train 4 and Train 5 do not have a delay change at timestamp $T+\Delta T$; the scheduled remaining running times of Train 5 is less than $\Delta T$ while that of Train 4 over $\Delta T$. In this case, Train 5 will never cause delay changes at the next timestamp $T+\Delta T$, but Train 4 is only suitable for some samples in which the delays are not changed between two adjacent timestamps. If we do not consider variable $R$, the prediction model cannot distinguish these two situations. Therefore, $R$ is also considered as an attribute for RT nodes.

These influencing factors are related to running trains, so they are considered attributes (features) of RT nodes. In summary, the attributes (features) of RT nodes are as follows:

$C$: the current delays of the predicted RT nodes;

$I$: the headways between the predicted RT nodes and their first forward RT nodes;

$S$: the scheduled running times of the predicted RT nodes from the current timestamp to the to-be-predicted timestamp (i.e., from timestamp $T$ to $T+\Delta T$);

$M$: the minimum running times of the predicted RT nodes from the current timestamp to the to-be-predicted timestamp (i.e., from timestamp $T$ to $T+\Delta T$);

$R$: the scheduled remaining running times of the predicted RT nodes from current timestamp to the to-be-predicted timestamp (i.e., from timestamp $T$ to $T+\Delta T$).

**(2) Features of TT nodes**

TT nodes correspond to the terminated trains which have already completed their itineraries when arriving at the terminal station in the current timestamp. Therefore, it is unnecessary to predict their delays in the next timestamp. However, as TT nodes may influence the rear predicted RT nodes, they must also be considered in this study. The current delays should be considered as the attributes of TT nodes. In addition, TT nodes do not have corresponding attributes $I$, $S$, $M$, and $R$ because terminated trains will not continue their itineraries. To distinguish from variable $C$ of the RT nodes, the current delay of the TT nodes is defined as $C^{TT}$.

**(3) Features of PS and TS nodes**

The other two types of nodes, namely PS and TS nodes, are stations/yards. We assume that the impact of stations/yards on trains depends mainly on the number of station tracks; therefore, the number of station tracks ($N$) is used as the feature of PS nodes. As described previously, the arrival-departure yards are treated as the TS nodes, so the number of station tracks is considered as the feature of TS nodes, labeled as $N^{TS}$. Although the features of PS and TS nodes are identical, they are different from the railway operation and management perspectives. For instance, trains only need to (not frequently) dwell at the passing (intermediate) stations, while there may be vehicle cleaning, inspection, and other tasks at terminal stations. We, therefore, assume that TS and PS nodes are different (even though they have the same features) from an operation and management perspective.

**(4) Features of Edges**

The edge *RT-rr-RT* connects consecutive trains (i.e., two RT nodes), to reflect the interactions of running trains. Because the *RT-rr-RT* edges reflect the influence of the forward train on the rear train, this edge is directed (e.g., RT 1 to RT 0 in **Fig. 4(A)**). Another edge *TT-tr-RT* reveals the influences of the forward terminated trains on the rear running trains. It is also directed (e.g., TT 1 to RT 6 in **Fig. 4(A)**), but the nodes have different attributes.

The *Station-ss-Station* edges reflect the physical structure of the railway network. The downstream station should influence the previous station, because we only focus on trains in one direction (i.e., the trains running to the terminal station); in other words, edges between TS (PS) and PS nodes are directed (e.g., PS 2 points to PS 1 and TS 1-1 points to PS 2 in **Fig. 4(A)**). However, arrival-departure yards of terminal stations interconnect, which allows trains in different yards to travel around. Hence, the edges between TS nodes are undirected (e.g., the edge between TS 1-1 and TS 1-2 in **Fig. 4(A)**).

The *Station-sr-RT* edge links the PS/TS nodes with RT nodes to uncover their relationship. *Station-sr-RT* should be the directed edge from the forward station to the rear train when the train is in the previous section or at the previous station. For instance, in **Fig. 1(A)**, Train 1 (i.e., RT 1) runs in the section (PS 1, PS 2) with PS 2 pointing to RT 1, while Train 5 (RT 5) dwells at PS 7, resulting in TS 1-3 pointing to RT 5.

Because GraphSAGE is not edge-weighted, the attributes of edges are equal to 1 when the nodes in the edges are linked. The node/edge types and features are summarized in **Table 3**.

**Table 3.** The summary of the types and attributes of nodes and edges

|  | Types | Brief introduction of the attributes |
|---|---|---|
| Node | RT node | $C$: the delays of the RT nodes at timestamp $T$. |
|  |  | $I$: the headways between the RT nodes and their first forward RT (TT) nodes. |
| | | *S*: the scheduled running time of the RT nodes from timestamp *T* to *T*+Δ*T*. |
| | | *M*: the minimum running times of the RT nodes from timestamp *T* to *T*+Δ*T* |
| | | *R*: the scheduled remaining running times of the RT nodes from timestamp *T* to *T*+Δ*T* |
| | TT node | $C^{TT}$: the delays of the TT nodes at timestamp *T* |
| | PS node | *N*: the number of station tracks |
| | TS node | $N^{TS}$: the number of station tracks at the corresponding yard |
| Edges (metapath) | *rr* (in *RT-rr-RT*) | 1 |
| | *tr* (in *TT-tr-RT*) | 1 |
| | *ss* (in *Station-ss-Station*) | 1 |
| | *sr* (in *Station-sr-RT*) | 1 |

## 5.3 Model Training

The mean absolute error (MAE), as defined by Eq. (4), was chosen as the loss function to train the SAGE-Het model.

$$MAE = \frac{1}{N} \sum_{i=1}^{N} \left| y_i - y_i' \right|, \tag{4}$$

where $y_i$ and $y_i'$ respectively correspond to the actual and predicted delays for the RT nodes, and *N* represents the total size of the dataset. The Adam optimizer with an initial learning rate of 0.001 was used as the model optimizer, and the global learning rate decreased by 10% when the decrease of the validation dataset over 10 epochs was less than 0.01. To prevent overfitting, the early-stop technique was also used during training. When there was less than a 0.01 decrease in the validation dataset over 30 epochs, the model stopped training, and the maximum number of epochs was set as 300. The data loader with a batch size of 64 was used.

The training techniques are exhibited in **Table 4**.

**Table 4**. A brief summarization of training techniques

| | |
|---|---|
| Optimizer | Adam |
| Loss function | MAE |
| Initial learning rate | 0.001 |
| Activation function | ReLU |
| Techniques to avoid over-fitting | 10 epochs with 10% learning rate reduction when less than 0.01 decrease |
| Techniques to decrease training time | 30 epochs less than 0.01 decrease |
| Mini-batch | 64 |
| Maximum epochs | 300 |

# 6. Result Analysis

In this section, we evaluate the proposed method. First, the comparative results between the proposed SAGE-Het and other existing methods are demonstrated. Then, we compare the the actual values with the predicted values based on SAGE-Het. Subsequently, we demonstrate the performance of the proposed model under different prediction horizons ($\Delta T$). Finally, we investigate the influence of train-train interactions on delay evolution.

*6.1 Compared with Existing Methods*

First, the proposed model is compared with the existing delay prediction models. The RF (Li et al., 2020b; Lulli et al., 2018; Nabian et al., 2019; Nair et al., 2019; Oneto et al., 2020), SVR (Barbour et al., 2018; Marković et al., 2015; Wang et al., 2021), and ANN (Peters et al., 2005; Yaghini et al., 2013) are some commonly-used methods in previous delay prediction studies, which also exhibit satisfactory performance. According to the *2018 RAS Competition* (Sciences, 2018), in the practical railway delay prediction process, some dispatchers assume that the delay is unchanged. In other words, once there is a delay, this delay is assumed to be the same until the next timestamp; thus, this prediction method was also considered and labeled as "Keep Constant". Each RT node serves as a sample for the baselines RF, SVR, and ANN. The input features of RT nodes for these baselines contain the variables $C$, $I$, $M$, $S$, and $R$ in **Table 2**. In addition, the delay of the first forward train of the RT node (labeled as $D$) is also considered to reflect the interaction between adjacent trains, and the number of station tracks at the station (i.e., the corresponding $N$ of station nodes) is also taken into account to represent the influence of station on the train. In summary, the input features of each RT node include $C$, $I$, $M$, S, $R$, $D$, and the $N/N^{TS}$ at the station. The proposed SAGE-Het cannot be compared with the existing HomoGNN-based delay prediction methods because different types of nodes are considered in this study, which HomoGNNs cannot handle.

The Introductions and related hyperparameter tuning of the baseline models are provided as follows.

**Keep Constant:** This method postulates that delays in the next timestamp are equal to the current timestamp, i.e., delays are kept constant until the train arrives at the terminal station.

**RF:** The Scikit-learn package (Pedregosa et al., 2011) was used to build the RF models in this study. The MSE is used as the loss function because the training time is extremely long with the MAE loss function. The hyperparameters, including *n_estimators* and *max_depth*, were tuned by GridSearchCV in the Scikit-learn package. The candidate values of *n_estimators* and *max_depth* were chosen through a grid search over [50,100,150,200] and [1, 2, 3, ..., 20], respectively.

**SVR**: The SVR model was also established by the Scikit-learn package. The hyperparameters $C$, *gamma*, and *epsilon* were optimized according to the validation dataset, with the candidate set [0.01,0.1,1,10].

**ANN:** The ANN model consisted of two hidden layers with 256 and 128 neurons, respectively. The activation function was ReLU, and the other training techniques were the same as those for SAGE-Het (described in Section 4.3).

In addition, some existing HetGNN algorithms, such as the Heterogeneous Graph Transformer (HGT) (Hu et al., 2020) and Heterogeneous Graph Attention Network (HAN) (Wang et al., 2019), were also selected for comparison with the proposed SAGE-based approach. Although they have not been applied in existing delay prediction studies, they have achieved great success in other prediction tasks (Mei et al., 2022; Yang et al., 2020).

**HAN:** This method combines the attention mechanism with the HetGNN model. In this study, the HGT model had four layers, each of which had 256 hidden units. The same training techniques as those used for SAGE-Het were applied to HAN.

**HGT:** This method merges the transformer architecture with the HetGNN model. The same hyperparameters and training techniques as those for HAN were used to train HGT.

The MAE and root-mean-square error (RMSE), as respectively given by Eqs. (4) and (5), were chosen as metrics to measure the performance on the testing dataset. All compared models were preprocessed in the same way as the SAGE-Het (e.g., normalized).

$$RMSE = \sqrt{\frac{1}{N} \sum_{i=1}^{N} (y_i - y_i')^2} \qquad (5)$$

The predictive performances measured by MAE of SAGE-Het and other baselines for all RT node samples are shown in **Fig. 5.** The results indicate that SAGE-Het achieved the best prediction performance at GZS-Net and CSS-Net. The Keep Constant method achieved the worst performance; this is unsurprising, as train operation is a dynamic process in which dispatchers adjust the scheduled operational plan of trains to reduce delays according to the real-time delay situation and timetable structure. Regarding the most commonly-used machine learning algorithms, namely the RF, SVR, and ANN, RF performed quite well in terms of the RMSE. Compared with RF regarding RMSE, SAGE-Het exhibited 1.6% and 6.2% improvement at GZS-Net and CSS-Net, respectively. However, SAGE-Het exhibited a comparative 18.6% and 17.2% improvement at GZS-Net and CSS-Net for MAE, respectively. The two existing HetGNNs, namely HAN and HGT, did not achieve better performance than SAGE-Het, which may have been because the HetGNN architecture does not consider the influences of too many entities, while a simpler model can achieve even better performance.

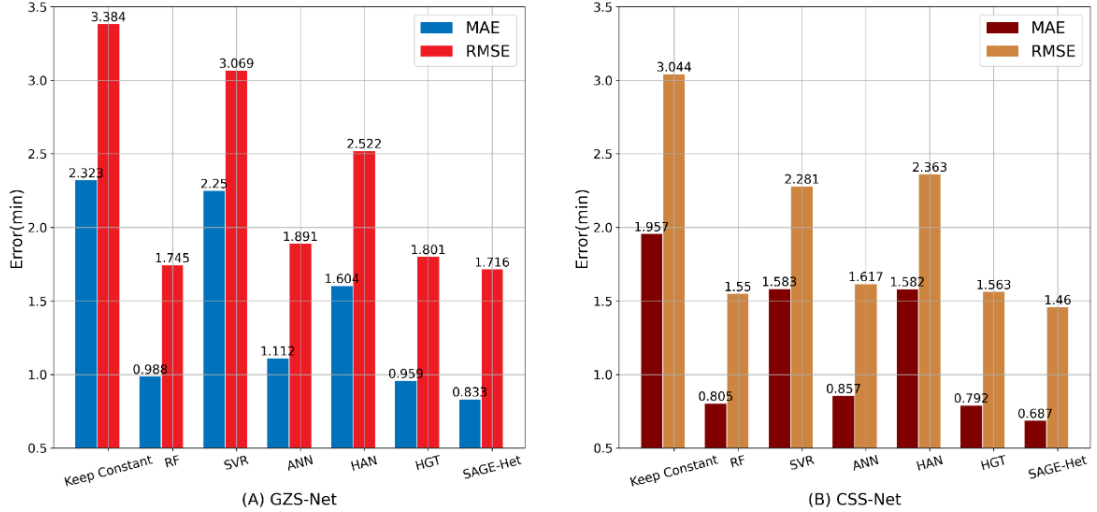

**Fig. 5.** The prediction performance of SAGE-Het and other baselines for all samples.

More attention should be given to delayed trains rather than early and on-time arrivals. Therefore, we investigate the predictive performance of SAGE-Het and other baselines for delayed RT node samples. The results in **Table 5** show that SAGE-Het outperforms other baselines, with better performance than RF, which has the best predictive performance in baseline models. In addition, the method Keep Constant has a more precise prediction performance for delayed RT nodes than for the whole samples. This is because early-arrival trains exist in the dataset. Early-arrival trains usually need to return to scheduled operation (by slowing down). However, delay recoveries are limited by the maximum recovery times, therefore, the change in delays of delayed trains is less significant than that of early-arrival trains. For instance, trains with a 5-minute early arrival may be on time at the next timestamp, while trains with a 5-minute delay may be delayed by 4 minutes if the maximum recovery time is 1 minute.

**Table 5**. The prediction performance of SAGE-Het and other baselines for delayed samples.

| Algorithms | GZS-MAE# | GZS-MAE# Increase (%) | GZS-RMSE# | GZS-RMSE# Increase (%) | CSS-MAE# | CSS-MAE# Increase (%) | CSS-RMSE# | CSS-RMSE# Increase (%) |
|---|---|---|---|---|---|---|---|---|
| Keep constant | 2.318 | 76% | 3.282 | 26.7%- | 1.77 | 55.1% | 3.022 | 29.4% |
| RF | 1.462 | 11% | 2.388 | 7.8% | 1.305 | 14.4% | 2.45 | 4.9% |
| SVR | 2.81 | 113.4% | 3.693 | 42.6% | 1.976 | 73.2% | 2.991 | 28% |
| ANN | 1.551 | 17.8% | 2.393 | 7.6% | 1.325 | 16.1% | 2.456 | 5.1% |
| HAN | 2.099 | 59.4% | 3.264 | 26% | 2.226 | 95.1% | 3.221 | 37.9% |
| HGT | 1.393 | 5.8% | 2.598 | 0.3% | 1.26 | 10.4% | 2.459 | 5.3% |
| SAGE-Het | 1.317 | - | 2.59 | - | 1.141 | - | 2.336 | - |

Note: the increase-related indicators are the increasing percentages of corresponding models compared with the SAGE-Het.

### 6.2 Comparison between the actual values and predicted values based on SAGE-Het

Then, to explore the prediction performance of SAGE-Het for different delay horizons, the predicted and actual delays are compared in **Fig. 6.** The results show that these scattered points were almost distributed near the 45-degree line, which demonstrates that the differences between the actual and predicted values were insignificant, i.e., SAGE-Het has high prediction accuracy. In addition, with the increment of actual delays, the prediction values did not deviate from the 45-degree line, which demonstrates that SAGE-Het has stable prediction performance for both short and long delays.

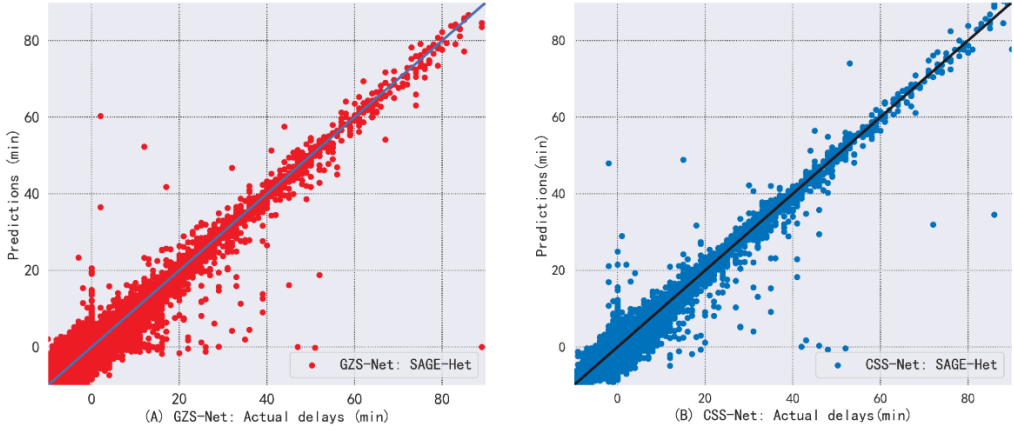

**Fig. 6.** The comparison of the actual and predicted values for SAGE-Het.

### 6.3 Performance for Different Prediction Time Intervals

The SAGE-Het was used to predict the delays 20 min later according to the current railway network delays above. To investigate the influence of different prediction time interval, the model performances for different prediction time interval (i.e., $\Delta T$ in Eq(1)), namely 10 and 30 min, were determined, and are exhibited in **Table 6**.

**Table 6**. The prediction performance under different prediction time intervals.

|  | GZS-Net: 10/20/30 (min) | | CSS-Net: 10/20/30 (min) | |
| --- | --- | --- | --- | --- |
|  | MAE | RMSE | MAE | RMSE |
| Keep Constant | 1.475/2.323/2.674 | 2.581/3.384/3.779 | 1.069/1.957/2.445 | 2.152/3.044/3.558 |
| RF | 0.654/0.988/1.055 | 1.315/1.745/2.051 | 0.540/0.805/0.926 | 1.2/1.550/1.818 |
| SVR | 1.484/2.308/2.827 | 2.589/3.064/3.799 | 1.069/1.508/1.838 | 2.157/2.245/2.811 |
| ANN | 0.688/1.112/1.241 | 1.339/1.891/2.231 | 0.553/0.857/0.966 | 1.278/1.617/1.867 |
| HAN | 1.101/1.604/1.771 | 1.781/2.522/2.897 | 1.101/1.582/1.721 | 2.015/2.363/2.606 |
| HGT | 0.657/0.959/1.011 | 1.3/1.801/1.897 | 0.546/0.792/0.883 | 1.308/1.563/1.697 |
| **SAGE-Het** | **0.558/0.833/0.901** | **1.201/1.716/1.897** | **0.447/0.687/0.781** | **1.236/1.460/1.666** |

Note: the bold fonts denote the best performance.

It can be seen from **Table 6** that the prediction performance decreased with the increase of the prediction time interval ($\Delta T$) at both GZS-Net and CSS-Net. In addition, SAGE-Het outperformed other baselines under different prediction time intervals. To better understand the prediction performance under different prediction time intervals, **Fig. 7** presents the residual distributions under different prediction time intervals. The results demonstrate that SAGE-Het achieved satisfactory performance for different prediction time intervals at the two railway networks. The residual distributions almost followed a bell-shaped distribution, which indicates that the residuals met both the zero-mean and normal distribution assumptions. In addition, the relationships between the cumulative distribution functions (CDFs) and the absolute values of residuals (|Residual|) are shown in **Fig. 8.** Each |Residual| value corresponds to the percentage of the absolute values of residuals that are no more than this value. For instance, when |Residual| equals 1 min, the corresponding CDF should be the percentage of the absolute values of residuals that are no more than 1 min. For both networks, when the prediction time interval was 10 min, the percentages of |Residual| no more than 1 min exceeded 80%, which indicates quite satisfactory performance. Even for the 30-min prediction time interval, the percentages of |Residual| no more than 1 min reached 70% for these two networks. The percentages of |Residual| no more than 3 min were over 90% for all three prediction time intervals (10, 20, 30 min) for the two networks.

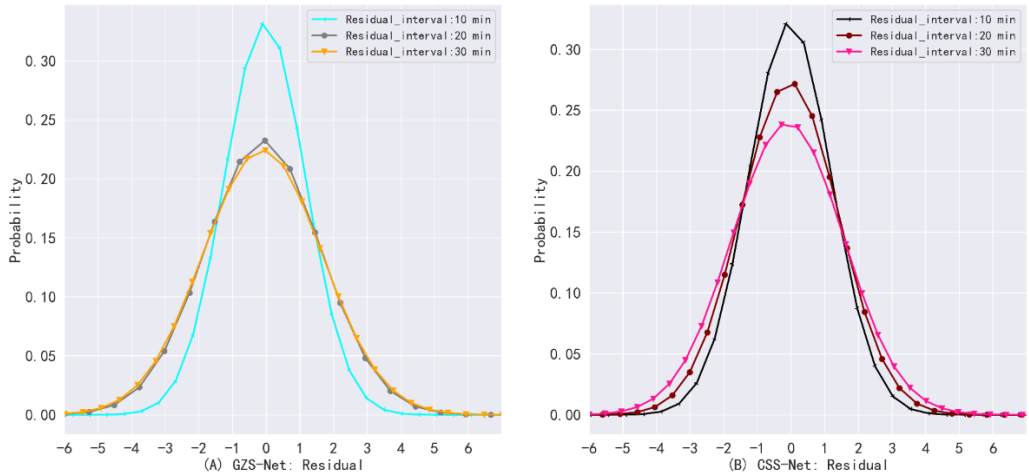

**Fig. 7.** The residual distributions for different prediction time intervals.

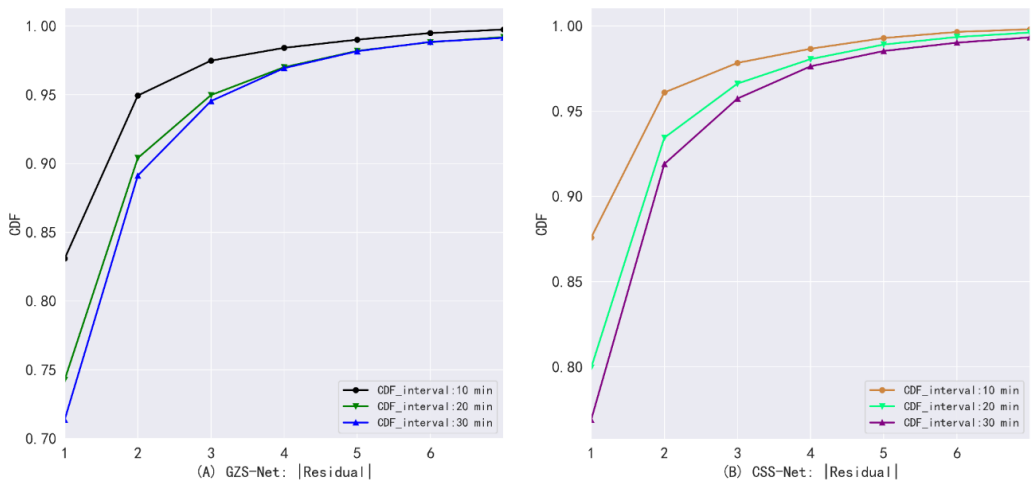

**Fig. 8.** The CDF distributions for different prediction time intervals.

*6.4 Influences of Train-Train Interaction*

In SAGE-Het, nodes are allowed to have different numbers of neighboring heterogeneous nodes, which are updated based on their neighboring nodes. Therefore, unlike traditional machine-learning approaches, SAGE-Het does not require all samples to be the same inputting dimension. In this way, the significance of interactions between nodes can be validated by canceling the corresponding connecting edges between nods. Since the interaction between adjacent trains is a crucial factor in the delay propagation process, the significance of train interactions on delay evolution can be tested by changing the linking between nodes. SAGE-Het considers the interactions between adjacent running trains and those between forward terminated trains and rear running trains by edges *RT-rr-RT* and *TT-tr-RT*. Therefore, firstly, we investigated whether these train interactions matter by canceling all edges between trains (i.e., deleting the edges *RT-rr-RT* and *TT-tr-RT*). In this way, the current states of running trains were only considered when updating the train-related edge (*RT-rr-RT* and *TT-tr-RT*) information, so this approach is labeled as **SAGE-Het-selflink**. In addition, Train headways also determine the train interactions. In other words, if the intervals between adjacent trains are large enough, the rear train has fewer possibilities to be influenced by the forward train (i.e., the train interaction does not matter in this situation). Therefore, the performances when canceling the edges whose headways between trains were more than a specified value (e.g., 20 min, 10 min, etc.) were investigated. For example, the method **SAGE-Het-cut-10** in **Table 7** denotes the deletion of the edges in the SAGE-Het model when the headways between consecutive running trains and those between the forward terminated trains and rear running trains were more than 10 min. Similarly, **SAGE-Het-cut-3**, **SAGE-Het-cut-5**, and **SAGE-Het-cut-20** were obtained.

**Table 7**. The prediction performance of SAGE-Het under different edge-linking methods.

| Algorithms | GZS-MAE | GZS-MAE Increases (%) | GZS-RMSE | GZS-RMSE Increases (%) | CSS-MAE | CSS-MAE Increases (%) | CSS-RMSE | CSS-RMSE Increases (%) |
|---|---|---|---|---|---|---|---|---|
| SAGE-Het-selflink | 0.965 | 15.8% | 1.876 | 9.3% | 0.741 | 7.9% | 1.607 | 10.1% |
| SAGE-Het-cut-3 | 0.925 | 11.0 % | 1.848 | 7.7% | 0.729 | 6.1% | 1.577 | 8.0% |
| SAGE-Het-cut-5 | 0.905 | 8.6% | 1.810 | 5.5% | 0.737 | 7.3% | 1.557 | 6.6% |
| SAGE-Het-cut-10 | 0.875 | 5.0% | 1.802 | 5.0% | 0.696 | 1.3% | 1.473 | 0.9% |
| SAGE-Het-cut-20 | 0.866 | 4.0% | 1.737 | 1.2% | 0.693 | 0.9% | 1.475 | 1.0% |
| SAGE-Het | 0.833 | - | 1.716 | - | 0.687 | - | 1.460 | - |

Note: the increase-related indicators are the increasing percentages of corresponding models compared with the SAGE-Het.

**Table 7** reveals that **SAGE-Het-selflink** achieved an obviously worse performance than SAGE-Het, indicating that train interactions are critical for delay propagation. Regarding different headways, the prediction performances of **SAGE-Het-cut-3** and **SAGE-Het-cut-5** were obviously decreased as compared with that of SAGE-Het. This is because the forward train will dramatically influence the rear train when their interval is so small that no (little) supplementary time can be utilized to maintain the safety distance. With the increase of the headways, the predictive performance was found to increase. The performance of **SAGE-Het-cut-20** was almost equal to that of the SAGE-Het model, which indicates that train interactions do not matter if the train interval is sufficient. In existing studies, train interaction was considered regardless of the length of the interval. To ensure consistency for all samples, the related information of forward trains must be inputted into the model. The proposed SAGE-Het model can modify the edge-linking method to address this shortcoming. While this study considered that all adjacent trains (stations) are linked, in future studies, the optimal edge-linking method can be learned by relying on some encoder blocks (Tygesen et al., 2022).

## 7. Conclusion and Discussions

This paper proposed a HetGNN model to investigate delay evolution in the railway network. In the proposed model, four kinds of nodes (e.g., RT, TT, PS, and TS nodes) are taken into account, and their interactions are represented by four edges (*RT-rr-RT*, *TT-tr-RT*, *Station-ss-Station*, and *Station-sr-RT*). A new neural architecture that is specific to the train delay prediction problem was developed; it is based on the GraphSAGE HomoGNN and HetGNN model, and we refer to it as SAGE-Het. The SAGE-Het model can process different types of nodes and capture the interactions between nodes. Based on the SAGE-Het model, the evolution of railway network delay was investigated by predicting the delays of running trains in future timestamps. The prediction performance of the SAGE-Het model was compared with that of existing delay prediction models (the RF, SVR, and ANN) and some advanced HetGNNs (HGT and HAN) that have achieved great success in other prediction tasks. Subsequently, the prediction performance under different prediction time intervals was compared. In addition, the prediction precision was demonstrated by calculating the residual distributions under diverse prediction time intervals. Lastly, the influence of train interactions on delay evolution was explored by changing the edge-linking methods in the SAGE-Het model. The main conclusions reached from these analyses are summarized as follows.

(1) The SAGE-Het model has better predictive performance than other existing delay prediction methods from the literature, as well as HGT, and HAN.

(2) The SAGE-Het model exhibits satisfactory predictive performance under different prediction time intervals. The absolute values of residuals that are no more than 1 min can reach up to 70%, even for predicting the delay 30 min into the future.

(3) Train interactions will influence the rear train when the intervals between adjacent trains are small, while this influence will disappear when the intervals are large enough, i.e., the forward trains will almost not influence the rear train when the train intervals are sufficient.

In comparison to the existing delay prediction methods, the proposed SAGE-Het approach can predict train delays more accurately, assisting railway operators and managers with more precise data and providing more comprehensive information to passengers. Once the prediction model based on SAGE-Het is trained, the dispatcher can callback the model to predict railway network delays whenever necessary. To callback the trained models requires only several (or few) seconds; for example, it only takes 3 seconds to load the well-trained SAGE-Het for test datasets with around 4600 graph samples in GZS-Net and CSS-Net. In addition, SAGE-Het, as a network-oriented approach, is capable of enhancing the comprehensive understanding of the delay evolution in railway networks, thus helping dispatchers make a comprehensive adjustment plan. Moreover, SAGE-Het enables nodes to connect with various numbers of heterogeneous nodes by edges, which can allow more flexible inputs and therefore does not require the missing features to be artificially filled. Interactions between nodes can also be explored by canceling corresponding edges based on SAGE-Het. For instance, the investigation of the impact of train-train interactions can improve understanding of the delay evolution, which can assist in obtaining a more applicable edge structure for our SAGE-Het in future work.

This research contains some shortcomings that need to be addressed in further studies. First, delays are updated at stations instead of continuously in this study, which may prevent the actual delays from being obtained in real time. Thus, the predicted delays will lag behind actual delays. However, once the real-time delay records are obtained, we can train a more precise prediction model based on SAGE-Het. In addition, this study only considered train and station attributes when establishing the SAGE-Het model to improve the prediction precision, although the proposed SAGE-Het model still achieved satisfactory performance. However, in the practical delay propagation process, railway network delay evolution results from the interactions of different network entities (e.g., weather, passenger flow, dispatching organization, etc.). Our proposed SAGE-Het model can theoretically gather the influences of all railway network entities. Therefore, in future studies, with the addition of more railway network entities to the SAGE-Het model and the selection of a more proper basic HomoGNN according to the edge characteristics, more precise delay prediction models can be constructed to reveal the evolution mechanism of railway network delay. In addition, in this study, the edges were determined by relying on delay propagation domain knowledge (e.g., adjacent trains are linked regardless of the interval length), while the investigation of the edge-linking method presented in Section 6.4 revealed that canceling some edges will not influence the results when the adjacent train intervals are sufficient. Therefore, in future research, some encoder blocks that can assist in learning the graph edge structure can be applied. Moreover, this HetGNN model can also be extended to other fields. For instance, in terms of the vehicle speed prediction task, the vehicles, pedestrians, and signal systems can be treated as different kinds of nodes, while corresponding edges can represent their relationships.

## Acknowledgment

This work was supported by the National Nature Science Foundation of China (grant numbers 71871188 and U1834209), the Research and development project of China National Railway Group Co., Ltd [grant number P2020X016], the Fundamental Research Funds for the Central Universities (grant number 2682021CX051), and the China Scholarship Council [grant number 202007000149]. We are grateful for the contributions made by our project partners